\documentclass[american]{llncs}
\usepackage[T1]{fontenc}
\usepackage[latin9]{inputenc}
\usepackage{array}
\usepackage{float}
\usepackage{booktabs}
\usepackage{slashed}
\usepackage{multirow}
\usepackage{amstext}
\usepackage{graphicx}

\makeatletter

\DeclareFontEncoding{LGR}{}{}
\DeclareTextSymbol{\~}{LGR}{126}
\newcommand{\lyxmathsym}[1]{\ifmmode\begingroup\def\b@ld{bold}
  \text{\ifx\math@version\b@ld\bfseries\fi#1}\endgroup\else#1\fi}

\titlerunning{AI Researchers, Games are Your Friends!}

\makeatother

\usepackage{babel}
\begin{document}

\title{AI Researchers, Video Games Are Your Friends!}

\author{Julian Togelius}

\institute{New York University\\
julian@togelius.com}
\maketitle
\begin{abstract}
If you are an artificial intelligence researcher, you should look to video games as ideal testbeds for the work you do. If you are a video game developer, you should look to AI for the technology that makes completely new types of games possible. This chapter lays out the case for both of these propositions. It asks the question ``what can video games do for AI'', and discusses how in particular general video game playing is the ideal testbed for artificial general intelligence research. It then asks the question ``what can AI do for video games'', and lays out a vision for what video games might look like if we had significantly more advanced AI at our disposal. 
The chapter is based on my keynote at IJCCI 2015, and is written in an attempt to be accessible to a broad audience.
\end{abstract}

\section{Introduction}

Video games and artificial intelligence are two of my favorite topics. Both as work and hobby. The great thing is that they go together so well: there is a great need for video games in artificial intelligence and for artificial intelligence in video games. In this chapter, I discuss what video games can do for AI and what AI can do for video games.

In the first part, I discuss the need for benchmarks in AI research and how games have historically been used as AI benchmarks. I then argue the advantages of video games over classic board games as AI benchmarks, and in particular the advantages of \emph{general} video game playing. I present the general video game playing competition and benchmark, and the vision of having games both generated and played automatically. I discuss how this fits into the idea of artificial general intelligence, the idea of developing AI that is good not only at a single thingsbut at all things, or at least most of them.

In the second part of the chapter, I discuss what AI can do in and for games. Lots of things, it turns out---playing them is what most people think of first, and it is true that there is a need for skilled and interesting adversaries and other non-player characters in many games---but perhaps even more exciting is all the possibilities that AI offers for modeling players, generating levels and perhaps even whole games, adapting games to suit players, and assisting game designers. The second section is structured as a vision of what playing an open-world game might be like in a future where we have the AI technologies to truly make the game we like, followed by a brief description of some of the research challenges involved in getting there.

It is important to note that this paper does not go into any technical depth on any particular topic, nor is it a comprehensive survey of the field. It is instead meant as an accessible, informal and inspirational introduction as well as a long-form argument. It is equal parts propaganda and science fiction. However, throughout the text I provide a number of references for further reading if you are interested in knowing the technical details or the full state of the field.

\section{What video games can do for AI}

The most important thing for humanity to do right now is to invent true artificial intelligence (AI): machines or software that can think and act independently in a wide variety of situations. Once we have artificial intelligence, it can help us solve all manner of other problems.

Luckily, thousands of researchers around work on inventing artificial intelligence. While most of them work on ways of using known AI algorithms to solve new or existing problems, some work on the overarching problem of artificial general intelligence. I do both. As I see it, addressing applied problems spur the invention of new algorithms, and the availability of new algorithms make it possible to address new problems. Having concrete problems to try to solve with AI is necessary in order to make progress; if you try to invent AI without having something to use it for, you will not know where to start. My chosen domain is games, and I will explain why this is the most relevant domain to work on if you are serious about AI.

But first, let us acknowledge that AI has gotten a lot of attention recently. In particular work on ``deep learning'' is being discussed in mainstream press as well as turned into startups that get bought by giant companies for bizarre amounts of money. There have been some very impressive advances during the past few years in identifying objects in images, understanding speech, matching names to faces, translating text and other such tasks. By some measures, the winner of the recent ImageNet contest is better than humans at correctly naming things in images~\cite{deng2009imagenet,krizhevsky2012imagenet}; sometimes I think Facebook's algorithms are better than I am at recognizing the faces of my acquaintances~\cite{taigman2014deepface}.

With few exceptions, the tasks that deep neural networks have excelled at are what are called pattern recognition problems~\cite{duda2001pattern}. Basically, take some large amount of data (an image, a song, a text) and output some other (typically smaller) data, such as a name, a category, another image or a text in another language. To learn to do this, they look at tons of data to find patterns. In other words, the neural networks are learning to do the same work as our brain's sensory systems: sight, hearing, touch and so on. To a lesser extent they can also do some of the job of our brain's language centra.

However, this is not all that intelligence is. We humans don't just sit around and watch things all day. We do things: solve problems by taking decisions and carrying them out. We move about and we manipulate our surroundings. (Sure, some days we stay in bed almost all day, but most of the rest of the time we are active in one way or another.) Our intelligence evolved to help us survive in a hostile environment, and doing that meant both reacting to the world and planning complicated sequences of actions, as well as adapting to changing circumstances~\cite{barkow96adapted,buss2015evolutionary}. Pattern recognition - identifying objects and faces, understanding speech and so on - is an important component of intelligence, but should really be thought of as one part of a complete system which is constantly working on figuring out what to do next. Trying to invent artificial intelligence while only focusing on pattern recognition is like trying to invent the car while only focusing on the wheels.

\subsection{The need for AI benchmarks}

In order to build a complete artificial intelligence we therefore need to build a system that takes actions in some kind of environment. How can we do this? Perhaps the most obvious idea is to embody artificial intelligence in robots. And indeed, robotics has shown us how even the most mundane tasks, such as walking in terrain or grabbing strangely shaped objects, are really rather hard to accomplish for robots~\cite{arkin1998behavior}. In the eighties, robotics research largely refocused on these kind of "simple" problems, which led to progress in applications as well as a better understanding of what intelligence is all about~\cite{brooks1991intelligence}. The last few decades of progress in robotics has fed into the development of self-driving cars, which is likely to become one of the areas where AI technology will revolutionize society in the near future.

Now, working with robots clearly has its downsides. Robots are expensive, complex and slow. When I started my PhD, my plan was to build robot software that would learn evolutionarily from its mistakes in order to develop increasingly complex and general intelligence---this undertaking generally goes by the name ``evolutionary robotics''~\cite{nolfi2000evolutionary}. But I soon realized that in order for my robots to learn from their experiences, they would have to attempt each task thousands of times, with each attempt maybe taking a few minutes. This meant that even a simple experiment would take several days - even if the robot would not break down (it usually would) or start behaving differently as the batteries depleted or motors warmed up. In order to learn any more complex intelligence I would have to build an excessively complex (and expensive) robot with advanced sensors and actuators, further increasing the risk of breakdown. I also would have to develop some very complex environments where complex skills could be learned. This all adds up, and quickly becomes unmanageable. Problems such as these is why the field of evolutionary robotics has not scaled up to evolve more complex intelligence.

I was too ambitious and impatient for that. I wanted to create complex intelligence that could learn from experience. So I turned to video games.

\subsection{Games as AI benchmarks}

Games and artificial intelligence have a long history together. Even since before artificial intelligence was recognized as a field, early pioneers of computer science wrote game-playing programs because they wanted to test whether computers could solve tasks that seemed to require "intelligence". Alan Turing, arguably the principal inventor of computer science, (re)invented the Minimax algorithm and used it to play Chess~\cite{turing1953digital}. (As no computer had been built yet, he performed the calculations himself using pen and paper.) Chess was for a long time one of the most important AI benchmarks~\cite{newell1958chess}. Arthur Samuel was the first to invent the form of machine learning that is now called reinforcement learning; he used it in a program that learned to play Checkers by playing against itself~\cite{samuel1959some}. Much later, IBM's Deep Blue computer famously won against the reigning grandmaster of Chess, Gary Kasparov, in a much-publicized 1997 event~\cite{campbell2002deep,newborn97kasparov}. Currently, many researchers around the world work on developing better software for playing the board game Go; up until recently, the best software is still no match for good human players~\cite{lee09computational,muller2002computer}. Between the first and the second revision of this chapter, Google DeepMind (Google's primary AI research division) announced in \emph{Nature} that their \emph{AlphaGo} Go-playing program had beaten the European champion at this game~\cite{silver2016mastering}.

Classic board game such as Chess, Checkers and Go are nice and easy to work with as they are very simple to model in code and can be simulated extremely fast - you could easily make millions of moves per second on a modern computer - which is indispensable for many AI techniques. Also, they seem to require thinking to play well. Many classib both depth and accessibility, meaning that they take ``a minute to learn, but a lifetime to master''. It is indeed the case that games have a lot to do with learning, and good games are able to constantly teach us more about how to play them. Indeed, to some extent the fun in playing a game consists in learning them and when there is nothing more to learn we largely stop enjoying them. This suggests that better-designed games are also better benchmarks for artificial intelligence. However, judging from the fact that now have (relatively simple) computer programs that can play Chess better than any human, it is clear that you don't need to be truly, generally intelligent to play such games well. When you think about it, they exercise only a very narrow range of human thinking skills; it's all about turn-based movements on a discrete grid of a few pieces with very well-defined, deterministic behavior.

But, despite what your grandfather might want you to believe, there's more to games than classical board games. In addition to all kinds of modern boundary-pushing board games, card games and role-playing games, there's also video games. Video games owe their massive popularity at least partly to that they engage multiple senses and multiple cognitive skills. Take a game such as Super Mario Bros. It requires you not only to have quick reactions, visual understanding and motoric coordination, but also to plan a path through the level, decide about tradeoffs between various path options (which include different risks and rewards), predict the future position and state of enemies and other characters of the level, predict the physics of your own running and jumping, and balance the demands of limited time to finish the level with multiple goals. Other games introduce demands of handling incomplete information (e.g. \emph{StarCraft}), understanding narrative (e.g. \emph{Skyrim}), or very long-term planning (e.g. \emph{Civilization}).

On top of this, video games run inside controllable environments in a computer and many (though not all) video games can be sped up to many times the original speed. It is simple and cheap to get started, and experiments can be run many thousands of times in quick succession, allowing the use of learning algorithms.

So it is not surprising that AI researchers are increasingly turning to video games as benchmarks. Researchers such as myself have adapted a number of video games to function as AI benchmarks. To make it easier to participate in this field and to provide common challenges for researchers to work on, we have organized competitions where researchers can submit their best game-playing AIs and test them against the best that other researchers can produce. Having recurring competitions based on the same game allows competitors to refine their approaches and methods, hoping to win next year. Games for which we have run such competitions include \emph{Super Mario Bros}~\cite{karakovskiy2012mario,togelius2013mario}, \emph{StarCraft}~\cite{ontanon2013survey}, the \emph{TORCS} racing game~\cite{loiacono20102009}, \emph{Ms. Pac-Man}~\cite{rohlfshagen2011ms}, a generic \emph{Street Fighter}-style figthing game~\cite{lu2013fighting}, Angry Birds~\cite{renz2015aibirds} and several others. In most of these competitions, we have seen performance of the winning AI player improve every time the competition is run. These competitions play an important role in catalyzing research in the community, and every year many papers are published where the competition software is used for benchmarking some new AI method. There are by now a set of best practices for how to organize such competition so as to maximize research value~\cite{togelius2014run}. Thus, we advance AI through game-based competitions.

\subsection{Artificial general intelligence and general game playing}

There's a problem with the picture I just painted. Can you spot it?

That's right. Game specificity. The problem is that improving how well an artificial intelligence plays a particular game is not necessarily helping us improve artificial intelligence in general. It's true that in most of the game-based competitions mentioned above we have seen the submitted AIs get better every time the competition ran. But in most cases, the improvements were not because of better AI algorithms, but because of even more ingenious ways of using these algorithms for the particular problems. Sometimes this meant relegating the AI to a more peripheral role. For example, in the car racing competition the first years were dominated by AIs that used evolutionary algorithms to train a neural network to keep the car on the track. In later years, most of the best submissions used hand-crafted "dumb" methods to keep the car on the track, but used learning algorithms to learn the shape of the track to adapt the driving~\cite{loiacono20102009}. This is a clever solution to a very specific engineering problem but says very little about intelligence in general.

In order to make sure that what such a competition measures is anything approaching actual intelligence, we need to recast the problem. To do this, it's a great idea to define what it is we want to measure: general intelligence. Shane Legg and Marcus Hutter have proposed a very useful definition of intelligence, which is roughly the average performance of an agent on all possible problems~\cite{legg2007universal}. (In their original formulation, each problem's contribution to the average is weighed by its simplicity, but let's disregard that for now.) Obviously, testing an AI on all possible problems is not an option, as there are infinitely many problems. But maybe we could test our AI on just a sizable number of diverse problems? For example on a number of different video games~\cite{schaul2011measuring}?

The first thing that comes to mind here is to just to take a bunch of existing games for some game console, preferably one that could be easily emulated and sped up to many times real time speed, and build an AI benchmark on them. This is what the Arcade Learning Environment (ALE) does~\cite{bellemare2012arcade}. ALE lets you test your AI on more than a hundred games released for 70s vintage \emph{Atari 2600} console. The AI agents get feeds of the screen at pixel level, and have to respond with a joystick command. ALE has been used in a number of experiments, including those by the original developers of the framework. Perhaps most famously, Google Deep Mind published a paper in \emph{Nature} last year showing how they could learn to play several of the games with superhuman skill using deep learning (Q-learning on a deep convolutional network)~\cite{mnih2015human}.

ALE is an excellent AI benchmark, but has a key limitation. The problem with using Atari 2600 games is that there is only a finite number of them, and developing new games is a tricky process. The Atari 2600 is notoriously hard to program, and the hardware limitations of the console tightly constrain what sort of games can be implemented. More importantly, all of the existing games are known and available to everyone. This makes it possible to tune your AI to each particular game. Not only to train your AI for each game (DeepMind's results depend on playing each individual game millions of times to train on it) but to tune your whole system to work better on the games you know you will train on.

Can we do better than this? Yes we can! If we want to approximate testing our AI on all possible problems, the best we can do is to test it on a number of unseen problems. That is, the designer of the AI should not know which problems it is being tested on before the test. At least, this was our reasoning when we designed the \emph{General Video Game Playing Competition}.

\subsection{General video game playing}

The General Video Game Playing Competition (GVGAI) allows anyone to submit their best AI players to a special server, which will then use them to play ten games that no-one (except the competition organizers) have seen before~\cite{perez20152014,perez2016general}. These games are of the type that you could find on home computers or in arcades in the early eighties; some of them are based on existing games such as \emph{Boulder Dash}, \emph{Pac-Man}, \emph{Space Invaders}, \emph{Sokoban} and \emph{Missile Command}. The winner of the competition is the AI that plays these unseen games best. Therefore, it is impossible for the creator of the AI to tune their software to any particular game. Around 60 games are currently available for training your AI on and 20 unseen games are available to test on; every iteration of the competition increases this number as the testing games from the previous iteration become available to train on, and new testing games are created.

Now, 60 games is not such a large number; where do we get new games from? To start with, all the games are programmed in something called the \emph{Video Game Description Language} (VGDL)~\cite{ebner2013towards,schaul2013video}. This is a simple language we designed to to be able to write games in a compact and human-readable way, a bit like how HTML is used to write web pages. The language is designed explicitly to be able to encode classical arcade games; this means that the games are all based on the movement of and interaction between sprites in two dimensions. This is how essentially all video games were designed before \emph{Wolfenstein 3D}, and quite a few games are still designed that way. In any case, the simplicity of this language makes it very easy to write new games, either from scratch or as variations on existing games. (Incidentally, as an offshoot of this project we are exploring the use of VGDL in a prototyping tool for game developers.)

\subsection{General video game generation}

Even if it's simple to write new games, that doesn't solve the fundamental problem that someone has to write them, and design them first. For the GVG-AI competition to reach its full potential as a test of general AI, we need an endless supply of new games. For this, we need to generate them. We need software that can produce new games at the press of a button, and these need to be good games that are not only playable but also require genuine skill to win. (As a side effect, such games are likely to be enjoyable for humans.)

I know, designing software that can design complete new games (that are also good in some sense) sounds quite hard. And it is. However, I and a couple of others have been working on this problem on and off for a couple of years, and I'm firmly convinced it is doable. Cameron Browne has already managed to build a complete generator for playable (and enjoyable) board games~\cite{browne2010evolutionary}, and several people including myself have attempted to automatically generate video games using different methods~\cite{nelson2007towards,togelius2008experiment,cook2011multi,zook2014automatic}, or just generating interesting variations of existing video games~\cite{isaksen2015discovering}. Some of our recent work has focused on generating simple VGDL games, and though we've had some success there is much left to do~\cite{nielsen2015general,nielsen2015towards}. Also, it is clearly possible to generate parts of games, such as game levels; there has been plenty of research within the last five years on procedural content generation - the automatic generation of game content~\cite{shaker2015procedural}. Researchers have demonstrated that methods such as evolutionary algorithms, planning and answer set programming can automatically create levels, maps, stories, items and geometry, and basically any other content type for games~\cite{togelius2011searchbased,smith2011answer}. Now, the research challenges are to make these methods general (so that they work for all games, not just for a particular game) and more comprehensive, so that they can generate all aspects of a game including the rules. Most of the generative methods include some form of simulation of the games that are being generated, suggesting that the problems of game playing and game generation are intricately connected and should be considered together whenever possible.

Once we have extended the General Video Game Playing Competition with automated game generation, we have a much better way of testing generic game-playing ability than we have ever had before. The software can of course also be used outside of the competition, providing a way to easily test the general intelligence of game-playing AI.

\subsection{What kind of AI will we need?}

So far we have only talked about how to best test or evaluate the general intelligence of a computer program, not how to best create one. Well, this post is about why video games are essential for inventing AI, and I think that I have explained that pretty well: they can be used to fairly and accurately benchmark AI. But for completeness, let us consider which are the most promising methods for creating AIs of this kind. As mentioned above, (deep) neural networks have recently attracted lots of attention because of some spectacular results in pattern recognition. I believe neural networks and similar pattern recognition methods will have an important role to play for evaluating game states and suggesting actions in various game states. In many cases, evolutionary algorithms are more suitable than gradient-based methods when training neural networks for games.

But intelligence can not only be pattern recognition. (This is for the same reason that behaviorism is not a complete account of human behavior: people don't just map stimuli to responses, sometimes they also think.) Intelligence must also incorporate some aspect of planning, where future sequences of actions can be played out in simulation before deciding what to do. Recently an algorithm called Monte Carlo Tree Search, which simulates the consequences of long sequences of actions by doing statistics of random actions, has worked wonders on the board game Go~\cite{browne2012survey}. It has also done very well on GVGAI. Another family of algorithms that has recently shown great promise on game planning tasks is rolling horizon evolution~\cite{perez2013rolling}. Here, evolutionary algorithms are used not for long-term learning, but for short-term action planning.

I think the next wave of advances in general video game-playing AIs will come from ingenious combinations of neural networks, evolution and tree search. (Case in point: Google's recent success on the game of Go stemmed from a combination of Monte Carlo Tree Search and two different types of neural networks~\cite{silver2016mastering}.) And from algorithms inspired by these methods. The important thing is that both pattern recognition and planning will be necessary in various different capacities. Of course, we cannot predict what will work well in the future (otherwise it wouldn't be called research), but I bet that exploring various combinations of these method will inspire the invention of the next generation of AI algorithms.

\subsection{The even bigger picture}

Now, you might object that this is a very limited view of intelligence and AI. What about text recognition, listening comprehension, storytelling, bodily coordination, irony and romance? Our game-playing AIs can't do any of this, no matter if it can play all the arcade games in the world perfectly. To this I say: \emph{Patience! One day.} None of these things are required for playing early arcade games, that is true. But as we master these games and move on to include other genres of games in our benchmark, such as role-playing games, adventure games, simulation games and social network games, many of these skills will be required to play well. As we gradually increase the diversity of games we include in our benchmark, we will also gradually increase the breadth of cognitive skills necessary to play well. Of course, our game-playing AIs will have to get more advanced to cope. Understanding language, images, stories, facial expression and humor will be necessary. And don't forget that closely coupled with the challenge of general video game playing is the challenge of general video game generation, where plenty of other types of intelligence will be necessary. I am convinced that video games (in general) challenges all forms of intelligence except perhaps those closely related to bodily movement, and therefore that video games (in general) are the best testbed for artificial intelligence. An AI that can play almost any video game and create a wide variety of video games is, by any reasonable standard, intelligent.

"But why, then, are not most AI researchers working on general video game playing and generation?"

To this I say: \emph{Patience! One day.}

This argument has become rather long and winding. Let me sum it up in a handy paragraph, so you remember what this was all about:

It is crucial for artificial intelligence research to have good testbeds. Games are excellent AI testbeds because they pose a wide variety of challenges and are highly engaging. But they are also simpler, cheaper and faster than robots, permitting a lot of research that is not practically possible with robotics. Board games have been used in AI research since the field started, but in the last decade more and more researchers have moved to video games because they offer more diverse and relevant challenges. (They are also more fun.) Competitions play a big role in this. But putting too much effort into AI for a single game has limited value for AI in general. Therefore we created the General Video Game Playing Competition and its associated software framework. This is meant to be the most complete game-based benchmark for general intelligence. AIs are evaluated on playing not a single video game, but on multiple games which the AI designer has not seen before. It is likely that the next breakthroughs in general video game playing will come from a combination of neural networks, evolutionary algorithms and Monte Carlo Tree Search. Coupled with the challenge of playing these games is the challenge of generating new games and new game content. The plan is to have an infinite supply of games to test AIs on. While playing and generating simple arcade games tests a large variety of cognitive capacities - more diverse than any other AI benchmark - we are not yet at the stage where we test all of intelligence. But there is no reason to think we would not get there, given the wide variety of intelligence that is needed to play and design modern video games.

It is now time to turn the perspective around a full radian, and ask not what video games can do for AI, but what AI can do for video games.

\section{What AI can do for video games}

Let's start in the here and now. The phrase ``game AI'' is usually understood as the artificial intelligence you find inside a video game, for example for controlling various non-player characters (NPCs). But is there really any AI in a typical video game? Depends on what you mean. The kind of AI that goes into most video games deals with pathfinding and expressing behaviors that were designed by human designers. The sort of AI that we work on in university research labs is often trying to achieve more ambitious goals, and therefore often not yet mature enough to use in an actual game. Alex Champandard, a prominent developer/researcher at the interface between academic and game-industrial AI, suggests that the "next giant leap of game AI is actually artificial intelligence"~\cite{graft2015artificial}. And there's indeed lots of things we could do in games if we only had the AI techniques to do it.

So let's step into the future, and assume that many of the various AI techniques we are working on at the moment have reached perfection, and we could make games that use them. In other words, let's imagine what games would be like if we had good enough AI for anything we wanted to do with AI in games. Imagine that you are playing a game of the future.

You are playing an "open world" game, something like \emph{Grand Theft Auto V} or \emph{Skyrim}. Instead of going straight to the next mission objective in the city you are in, you decide to drive (or ride) five hours in some randomly chosen direction. The game makes up the landscape as you go along, and you end up in a new city that no human player has visited before. In this city, you can enter any house (though you might have to pick a few locks), talk to everyone you meet, and involve yourself in a completely new set of intrigues and carry out new missions. If you would have gone in a different direction, you would have reached a different city with different architecture, different people and different missions. Or a huge forest with realistic animals and eremites, or a secret research lab, or whatever the game comes up with.

Talking to these people you find in the new city is as easy as just talking to the screen. The characters respond to you in natural language that takes into account what you just said. These lines are not read by an actor but generated in real-time by the game. You could also communicate with the game though waving your hands around, dancing or using other exotic modalities for expressing emotions and intentions. Of course, in many (most?) cases you are still pushing buttons on a keyboard or controller, as that is often the most efficient way of telling the game what you want to do.

Perhaps needless to say, but all the non-player characters (NPCs) navigate and generally behave in a thoroughly believable way. For example, they will not get stuck running into walls or repeat the same sentence over and over (well, not more than an ordinary human would). This also means that you have interesting adversaries and collaborators to play any game with, without having to resort either to waiting for your friends to come online or to being matched with annoying thirteen year-olds.

Within the open world game, there are other games to play, for example by accessing virtual game consoles within the game or proposing to play a game with some NPC. These NPCs are capable of playing the various sub-games at whatever level of proficiency that fits with the game fiction, and they play with human-like playing styles. It is also possible to play the core game at different resolutions, for example as a management game or as a game involving the control of individual body parts, by zooming in or out. Whatever rules, mechanics and content are necessary to play these sub-games or derived games are invented by the game engine on the spot. Any of these games can be lifted out of the main game and played on its own.

The game senses how you feel while playing the game, and figures out which aspects of it you are good at as well as which parts you like (and conversely, which parts you suck at and despise). Based on this, the game constantly adapts itself to be more to your liking, for example by giving you more story, challenges and experiences that you will like in that new city which you reached by driving five hours in a randomly chosen direction. Or perhaps by changing its own rules. It's not just that the game is giving you more of what you already liked and mastered. Rather more sophisticatedly, the game models what you preferred in the past, and creates new content that answers to your evolving skills and preferences as you keep playing.

Although the game you are playing is endless, of infinite resolution and continuously adapts to your changing tastes and capabilities, you might still want to play something else at some point. So why not design and make your own game? Maybe because it's hard and requires lots of work? Sure, it's true that back in 2015 it required hundreds of people working for years to make a high profile game, and it required at least a handful of highly skilled professionals to make any notable game at all, even if small. But now that it's the future and we have advanced AI, this can be used not only inside of the game but also in the game design and development and process. So you simply switch the game engine to edit mode and start sketching a game idea. A bit of a storyline here, a character there, some mechanics over here and a set piece on top of it. The game engine immediately fills in the missing parts and provides you with a complete, playable game. Some of it is suggestions: if you have sketched an in-game economy but have no money sink, the game engine will suggest one for you, and if you have designed gaps that the player character can not jump over, the game engine will suggest changes to the gaps or to the jump mechanic. You can continue sketching, and the game engine will convert your sketches into details, or jump right in and start modifying the details of the game; whatever you do, the game engine will work with you to flesh out your ideas into a complete game with art, levels and characters. At any time you can jump in and play the game yourself, and you can also watch a number of artificial players play various parts of the game, including players that play like you would have played the game or like your friends (with different tastes and skills) would have played it.

If you ask me, I'd say that this is a rather enticing vision of the future. I'll certainly play a lot of games if this is what games will look like in a decade or so. But will they? Will we have the AI techniques to make all this possible? Well, me and a bunch of other people in the CI/AI in Games research community are certainly working on it. (Whether that means that progress is more or less likely to happen is another question...) My team and I are in some form working on all of the things discussed above, except the natural interaction parts (talking to the game etc).

Let's start with the goal of generating complete games~\cite{togelius2008experiment,togelius2014characteristics,nielsen2015towards,font2013towards}. This requires generating a large number of different aspects of the game, including levels, rules, items, quests, textures etc. The generation of various types of game content is commonly referred to as \emph{procedural content generation}~\cite{shaker2015procedural,togelius2013procedural}. We work mainly within the search-based procedural content generation paradigm~\cite{togelius2011searchbased}, where evolutionary algorithms are used to generate content; often, this takes the form of searching for game content that, according to a player model, creates some particular type of player experience~\cite{yannakakis2011experience}. This of course requires us to have models of player experience and player behavior~\cite{yannakakis2013player,smith2011inclusive,pedersen2010modeling,mahlmann2010predicting}, so we can predict what players will do when faced with a particular type of game content and how they will experience it. Given that we for the foreseeable future will not be able to completely automate all parts of the game creation process we need to find ways to involve humans inside the game and content generation process; we need \emph{mixed-initiative} tools that combine the best of human and machine creativity~\cite{liapis2013sentient,yannakakis2014mixed,shaker2013evolving,shaker2013ropossum}. In order to assess the quality of games and game content we need to be able to playtest them. Therefore we need strong AI capable of playing any game---which, not coincidentally, is what the first part of this chapter focuses on. Once you have a strong game-playing AI, you might also need to restrict it or otherwise modify it so that it plays the game in a human-like manner; it is common that strong AI players play in a somewhat ``machine-like way''~\cite{hingston2012believable,shaker2013turing,ortega2013imitating}.

By now you probably see how it all fits together. In order to generate games you need to generate various types of content, and in order to do that you need good player models and good artificial players to play the games in a human-like manners. But in order to develop good game-playing AI you need to test your players on multiple games, and in order to do so you need to automatically generate games and game content of high quality~\cite{yannakakis2014panorama}. It's like a web, where every part is dependent on every other part. Games are essential to furthering AI, but AI also has a lot to give games. This chapter has tried to explain some of the various ways in which these research questions interact.

This chapter is also an invitation to you to start working within the field of AI in games, and address some of its many fascinating questions. If you are already an AI researcher, you should consider working on games. If you are a researcher in a different field interested in games, consider taking the artificial intelligence perspective on the research problems associated with games. There is a lot of work to do, and you are welcome to join our research community.


\bibliographystyle{splncs03}
\bibliography{references}

\begin{thebibliography}{10}
\providecommand{\url}[1]{\texttt{#1}}
\providecommand{\urlprefix}{URL }

\bibitem{arkin1998behavior}
Arkin, R.: Behavior-based robotics. The MIT Press (1998)

\bibitem{barkow96adapted}
Barkow, J.H., Cosmides, L., Tooby, J.: The Adapted Mind: Evolutionary
  Psychology and the Generation of Culture. Oxford University Press (1996)

\bibitem{bellemare2012arcade}
Bellemare, M., Naddaf, Y., Veness, J., Bowling, M.: The arcade learning
  environment: An evaluation platform for general agents. Arxiv preprint
  arXiv:1207.4708  (2012)

\bibitem{brooks1991intelligence}
Brooks, R.: Intelligence without representation. Artificial Intelligence  47,
  139--159 (1991)

\bibitem{browne2012survey}
Browne, C., Powley, E., Whitehouse, D., Lucas, S., Cowling, P., Rohlfshagen,
  P., Tavener, S., Perez, D., Samothrakis, S., Colton, S.: {A Survey of Monte
  Carlo Tree Search Methods}. IEEE Transactions on Computational Intelligence
  and AI in Games  4:1,  1--43 (2012)

\bibitem{browne2010evolutionary}
Browne, C., Maire, F.: Evolutionary game design. Computational Intelligence and
  AI in Games, IEEE Transactions on  2(1),  1--16 (2010)

\bibitem{buss2015evolutionary}
Buss, D.: Evolutionary psychology: The new science of the mind. Psychology
  Press (2015)

\bibitem{campbell2002deep}
Campbell, M., Hoane, A.J., Hsu, F.h.: Deep blue. Artificial intelligence
  134(1),  57--83 (2002)

\bibitem{cook2011multi}
Cook, M., Colton, S.: Multi-faceted evolution of simple arcade games. In: IEEE
  Conference on Computational Intelligence in Games. pp. 289--296 (2011)

\bibitem{deng2009imagenet}
Deng, J., Dong, W., Socher, R., Li, L.J., Li, K., Fei-Fei, L.: Imagenet: A
  large-scale hierarchical image database. In: Computer Vision and Pattern
  Recognition, 2009. CVPR 2009. IEEE Conference on. pp. 248--255. IEEE (2009)

\bibitem{duda2001pattern}
Duda, R.O., Hart, P.E., Stork, D.G.: Pattern Classification. John Wiley \&
  Sons, New York, NY, 2 edn. (2001)

\bibitem{ebner2013towards}
Ebner, M., Levine, J., Lucas, S.M., Schaul, T., Thompson, T., Togelius, J.:
  Towards a video game description language  (2013)

\bibitem{font2013towards}
Font~Fern{\'a}ndez, J.M., Manrique~Gamo, D., Mahlmann, T., Togelius, J.:
  Towards the automatic generation of card games through grammar-guided genetic
  programming. In: EvoApps (2013)

\bibitem{graft2015artificial}
Graft, K.: When artificial intelligence in video games becomes... artificially
  intelligent. Gamasutra  (2015)

\bibitem{hingston2012believable}
Hingston, P.: Believable Bots. Springer (2012)

\bibitem{isaksen2015discovering}
Isaksen, A., Gopstein, D., Togelius, J., Nealen, A.: Discovering unique game
  variants. In: Computational Creativity and Games Workshop at the 2015
  International Conference on Computational Creativity (2015)

\bibitem{karakovskiy2012mario}
Karakovskiy, S., Togelius, J.: The mario ai benchmark and competitions.
  Computational Intelligence and AI in Games, IEEE Transactions on  4(1),
  55--67 (2012)

\bibitem{krizhevsky2012imagenet}
Krizhevsky, A., Sutskever, I., Hinton, G.E.: Imagenet classification with deep
  convolutional neural networks. In: Advances in neural information processing
  systems. pp. 1097--1105 (2012)

\bibitem{lee09computational}
Lee, C.S., Wang, M.H., Chaslot, G., Hoock, J.B., Rimmel, A., Teytaud, O., Tsai,
  S.R., Hsu, S.C., Hong, T.P.: {The Computational Intelligence of MoGo Revealed
  in Taiwan's Computer Go Tournaments}. IEEE Trans. Comput. Intellig. and AI in
  Games  1(1),  73--89 (2009)

\bibitem{legg2007universal}
Legg, S., Hutter, M.: {Universal Intelligence : A Definition of Machine
  Intelligence}. Minds and Machines  17(4),  391--444 (2007)

\bibitem{liapis2013sentient}
Liapis, A., Yannakakis, G.N., Togelius, J.: Sentient sketchbook: Computer-aided
  game level authoring. In: FDG. pp. 213--220 (2013)

\bibitem{loiacono20102009}
Loiacono, D., Lanzi, P.L., Togelius, J., Onieva, E., Pelta, D.A., Butz, M.V.,
  Lonneker, T.D., Cardamone, L., Perez, D., S{\'a}ez, Y., et~al.: The 2009
  simulated car racing championship. Computational Intelligence and AI in
  Games, IEEE Transactions on  2(2),  131--147 (2010)

\bibitem{lu2013fighting}
Lu, F., Yamamoto, K., Nomura, L.H., Mizuno, S., Lee, Y., Thawonmas, R.:
  Fighting game artificial intelligence competition platform. In: Consumer
  Electronics (GCCE), 2013 IEEE 2nd Global Conference on. pp. 320--323. IEEE
  (2013)

\bibitem{mahlmann2010predicting}
Mahlmann, T., Drachen, A., Togelius, J., Canossa, A., Yannakakis, G.N.:
  Predicting player behavior in tomb raider: Underworld. In: Computational
  Intelligence and Games (CIG), 2010 IEEE Symposium on. pp. 178--185. IEEE
  (2010)

\bibitem{mnih2015human}
Mnih, V., Kavukcuoglu, K., Silver, D., Rusu, A.A., Veness, J., Bellemare, M.G.,
  Graves, A., Riedmiller, M., Fidjeland, A.K., Ostrovski, G., et~al.:
  Human-level control through deep reinforcement learning. Nature  518(7540),
  529--533 (2015)

\bibitem{muller2002computer}
M{\"u}ller, M.: Computer go. Artificial Intelligence  134(1),  145--179 (2002)

\bibitem{nelson2007towards}
Nelson, M., Mateas, M.: Towards automated game design. In: Procedings of the
  10th Congress of the Italian Association for Artificial Intelligence (2007)

\bibitem{newborn97kasparov}
Newborn, M.: Kasparov Vs. Deep Blue: Computer Chess Comes of Age. Springer
  (1997)

\bibitem{newell1958chess}
Newell, A., Shaw, J.C., Simon, H.A.: Chess-playing programs and the problem of
  complexity. IBM Journal of Research and Development  2(4),  320--335 (1958)

\bibitem{nielsen2015general}
Nielsen, T.S., Barros, G., Togelius, J., Nelson, M.J.: General video game
  evaluation using relative algorithm performance profiles. In: Proceedings of
  EvoApplications (2015)

\bibitem{nielsen2015towards}
Nielsen, T.S., Barros, G.A., Togelius, J., Nelson, M.J.: Towards generating
  arcade game rules with vgdl. In: IEEE Conference on Computational
  Intelligence in Games (2015)

\bibitem{nolfi2000evolutionary}
Nolfi, S., Floreano, D.: Evolutionary robotics. MIT Press, Cambridge, MA (2000)

\bibitem{ontanon2013survey}
Ontan{\'o}n, S., Synnaeve, G., Uriarte, A., Richoux, F., Churchill, D., Preuss,
  M.: A survey of real-time strategy game ai research and competition in
  starcraft. Computational Intelligence and AI in Games, IEEE Transactions on
  5(4),  293--311 (2013)

\bibitem{ortega2013imitating}
Ortega, J., Shaker, N., Togelius, J., Yannakakis, G.N.: Imitating human playing
  styles in super mario bros. Entertainment Computing  4(2),  93--104 (2013)

\bibitem{pedersen2010modeling}
Pedersen, C., Togelius, J., Yannakakis, G.N.: Modeling player experience for
  content creation. IEEE Transactions on Computational Intelligence and AI in
  Games  2(1),  54--67 (2010)

\bibitem{perez2013rolling}
Perez, D., Samothrakis, S., Lucas, S., Rohlfshagen, P.: Rolling horizon
  evolution versus tree search for navigation in single-player real-time games.
  In: Proceedings of the 15th annual conference on Genetic and evolutionary
  computation. pp. 351--358. ACM (2013)

\bibitem{perez20152014}
Perez, D., Samothrakis, S., Togelius, J., Schaul, T., Lucas, S., Cou{\"e}toux,
  A., Lee, J., Lim, C.U., Thompson, T.: The 2014 general video game playing
  competition. IEEE Transactions on Computational Intelligence and AI in Games
  (2015)

\bibitem{perez2016general}
Perez-Liebana, D., Samothrakis, S., Togelius, J., Schaul, T., Lucas, S.M.:
  General video game ai: Competition, challenges and opportunities. In: AAAI
  (2016)

\bibitem{renz2015aibirds}
Renz, J.: Aibirds: The angry birds artificial intelligence competition. In:
  AAAI. pp. 4326--4327 (2015)

\bibitem{rohlfshagen2011ms}
Rohlfshagen, P., Lucas, S.M.: Ms pac-man versus ghost team cec 2011
  competition. In: Evolutionary Computation (CEC), 2011 IEEE Congress on. pp.
  70--77. IEEE (2011)

\bibitem{samuel1959some}
Samuel, A.: Some studies in machine learning using the game of checkers. IBM
  Journal of Research and Development  3(3),  210--229 (1959)

\bibitem{schaul2011measuring}
Schaul, T., Togelius, J., Schmidhuber, J.: Measuring intelligence through
  games. Arxiv preprint arXiv:1109.1314  (2011)

\bibitem{schaul2013video}
Schaul, T.: A video game description language for model-based or interactive
  learning. In: Proceedings of the IEEE Conference on Computational
  Intelligence in Games. IEEE Press, Niagara Falls (2013)

\bibitem{shaker2013evolving}
Shaker, N., Shaker, M., Togelius, J.: Evolving playable content for cut the
  rope through a simulation-based approach. In: AIIDE (2013)

\bibitem{shaker2013ropossum}
Shaker, N., Shaker, M., Togelius, J.: Ropossum: An authoring tool for
  designing, optimizing and solving cut the rope levels. In: AIIDE (2013)

\bibitem{shaker2015procedural}
Shaker, N., Togelius, J., Nelson, M.J.: Procedural content generation in games:
  A textbook and an overview of current research. Procedural Content Generation
  in Games: A Textbook and an Overview of Current Research  (2015)

\bibitem{shaker2013turing}
Shaker, N., Togelius, J., Yannakakis, G.N., Poovanna, L., Ethiraj, V.S.,
  Johansson, S.J., Reynolds, R.G., Heether, L.K., Schumann, T., Gallagher, M.:
  The turing test track of the 2012 mario ai championship: entries and
  evaluation. In: IEEE Conference on Computational Intelligence in Games (CIG).
  pp. 1--8. IEEE (2013)

\bibitem{silver2016mastering}
Silver, D., Huang, A., Maddison, C.J., Guez, A., Sifre, L., van~den Driessche,
  G., Schrittwieser, J., Antonoglou, I., Panneershelvam, V., Lanctot, M.,
  et~al.: Mastering the game of go with deep neural networks and tree search.
  Nature  529(7587),  484--489 (2016)

\bibitem{smith2011inclusive}
Smith, A.M., Lewis, C., Hullett, K., Smith, G., Sullivan, A.: An inclusive
  taxonomy of player modeling. University of California, Santa Cruz, Tech. Rep.
  UCSC-SOE-11-13  (2011)

\bibitem{smith2011answer}
Smith, A.M., Mateas, M.: Answer set programming for procedural content
  generation: A design space approach. Computational Intelligence and AI in
  Games, IEEE Transactions on  3(3),  187--200 (2011)

\bibitem{taigman2014deepface}
Taigman, Y., Yang, M., Ranzato, M., Wolf, L.: Deepface: Closing the gap to
  human-level performance in face verification. In: Proceedings of the IEEE
  Conference on Computer Vision and Pattern Recognition. pp. 1701--1708 (2014)

\bibitem{togelius2011searchbased}
Togelius, J., Yannakakis, G., Stanley, K., Browne, C.: Search-based procedural
  content generation: A taxonomy and survey. IEEE Transactions on Computational
  Intelligence and AI in Games (99) (2011)

\bibitem{togelius2014run}
Togelius, J.: How to run a successful game-based ai competition. IEEE
  Transactions on Computational Intelligence and AI in Games  (2014)

\bibitem{togelius2013procedural}
Togelius, J., Champandard, A.J., Lanzi, P.L., Mateas, M., Paiva, A., Preuss,
  M., Stanley, K.O.: Procedural content generation: Goals, challenges and
  actionable steps. Dagstuhl Follow-Ups  6 (2013)

\bibitem{togelius2014characteristics}
Togelius, J., Nelson, M.J., Liapis, A.: Characteristics of generatable games.
  In: Foundations of Digital Games. vol.~9, p.~20 (2014)

\bibitem{togelius2008experiment}
Togelius, J., Schmidhuber, J.: {An Experiment in Automatic Game Design}. In:
  Proceedings of the IEEE Symposium on Computational Intelligence and Games
  (2008)

\bibitem{togelius2013mario}
Togelius, J., Shaker, N., Karakovskiy, S., Yannakakis, G.N.: {The Mario AI
  Championship 2009-2012}. AI Magazine  34(3),  89--92 (2013)

\bibitem{turing1953digital}
Turing, A.M., Bates, M., Bowden, B., Strachey, C.: Digital computers applied to
  games. Faster than thought  101 (1953)

\bibitem{yannakakis2014mixed}
Yannakakis, G.N., Liapis, A., Alexopoulos, C.: Mixed-initiative co-creativity.
  In: Proceedings of the 9th Conference on the Foundations of Digital Games
  (2014)

\bibitem{yannakakis2013player}
Yannakakis, G.N., Spronck, P., Loiacono, D., Andr{\'e}, E.: Player modeling.
  Dagstuhl Follow-Ups  6 (2013)

\bibitem{yannakakis2011experience}
Yannakakis, G.N., Togelius, J.: Experience-driven procedural content
  generation. Affective Computing, IEEE Transactions on  2(3),  147--161 (2011)

\bibitem{yannakakis2014panorama}
Yannakakis, G.N., Togelius, J.: A panorama of artificial and computational
  intelligence in games. IEEE Transactions on Computational Intelligence and AI
  in Games  (2014)

\bibitem{zook2014automatic}
Zook, A., Riedl, M.O.: Automatic game design via mechanic generation. In: AAAI.
  pp. 530--537 (2014)

\end{thebibliography}

\end{document}